\documentclass[runningheads]{llncs}
\usepackage{graphicx}
\usepackage{tikz}
\usepackage{comment}
\usepackage{amsmath,amssymb} % define this before the line numbering.
\usepackage{color}

\usepackage{enumerate}
\usepackage{enumitem}
\usepackage{multirow}
\usepackage{caption, subcaption}
\usepackage{booktabs}
\usepackage{diagbox}
\usepackage{array}
\usepackage{color}

\usepackage{bbding}

\usepackage{cite}

\newcommand{\etal}{\textit{et al}.}
\newcommand{\ie}{\textit{i}.\textit{e}.}
\newcommand{\eg}{\textit{e}.\textit{g}.}

\makeatletter
\def\@fnsymbol#1{\ensuremath{\ifcase#1\or *\or \dagger\or \ddagger\or
   \mathsection\or \mathparagraph\or \|\or **\or \dagger\dagger
   \or \ddagger\ddagger \else\@ctrerr\fi}}

\begin{document}
% \renewcommand\thelinenumber{\color[rgb]{0.2,0.5,0.8}\normalfont\sffamily\scriptsize\arabic{linenumber}\color[rgb]{0,0,0}}
% \renewcommand\makeLineNumber {\hss\thelinenumber\ \hspace{6mm} \rlap{\hskip\textwidth\ \hspace{6.5mm}\thelinenumber}}
% \linenumbers
\pagestyle{headings}
\mainmatter
\def\ECCVSubNumber{5068}  % Insert your submission number here

\title{Context-Aware RCNN: A Baseline for Action Detection in Videos} % Replace with your title

% INITIAL SUBMISSION
% \begin{comment}
% \titlerunning{ECCV-20 submission ID \ECCVSubNumber}
% \authorrunning{ECCV-20 submission ID \ECCVSubNumber}
% \author{Anonymous ECCV submission}
% \institute{Paper ID \ECCVSubNumber}
%\end{comment}
%******************

% CAMERA READY SUBMISSION
% \begin{comment}
\titlerunning{Context-Aware RCNN: A Baseline for Action Detection in Videos}
% If the paper title is too long for the running head, you can set
% an abbreviated paper title here
%

\author{
Jianchao Wu\inst{1}\thanks{Part of the work is done during an internship at SenseTime EIG Research.} \and
Zhanghui Kuang\inst{2} \and
Limin Wang\inst{1}\thanks{Corresponding author (lmwang@nju.edu.cn).} \and
Wayne Zhang\inst{2}\and
Gangshan Wu\inst{1}
}

\authorrunning{J. Wu et al.}
% \authorrunning{J. Wu, Z. Kuang, L. Wang, W. Zhang, and G. Wu}
% First names are abbreviated in the running head.
% If there are more than two authors, 'et al.' is used.
%
\institute{State Key Laboratory for Novel Software Technology, Nanjing University, China \and
SenseTime Research
}

% \end{comment}
%******************
\maketitle

\begin{abstract}
Video action detection approaches usually conduct actor-centric action recognition over RoI-pooled features following the standard pipeline of Faster-RCNN.
In this work, we first empirically find the recognition accuracy is highly correlated with the bounding box size of an actor, and thus higher resolution of actors contributes to better performance.
However, video models require dense sampling in time to achieve accurate recognition.
To fit in GPU memory, the frames to backbone network must be kept low-resolution, resulting in a coarse feature map in RoI-Pooling layer.
Thus, we revisit RCNN for actor-centric action recognition via cropping and resizing image patches around actors before feature extraction with I3D deep network.
Moreover, we found that expanding actor bounding boxes slightly and fusing the context features can further boost the performance.
Consequently, we develop a surpringly effective baseline (Context-Aware RCNN) and it achieves new state-of-the-art results on two challenging action detection benchmarks of AVA and JHMDB.
Our observations challenge the conventional wisdom of RoI-Pooling based pipeline and encourage researchers rethink the importance of resolution in actor-centric action recognition. Our approach can serve as a strong baseline for video action detection and is expected to inspire new ideas for this filed.
The code is available at \url{https://github.com/MCG-NJU/CRCNN-Action}.
\keywords{Action Detection, Context-aware RCNN, Baseline}
\end{abstract}

\section{Introduction}

\begin{figure}
	\begin{center}
		\includegraphics[width=0.6\linewidth]{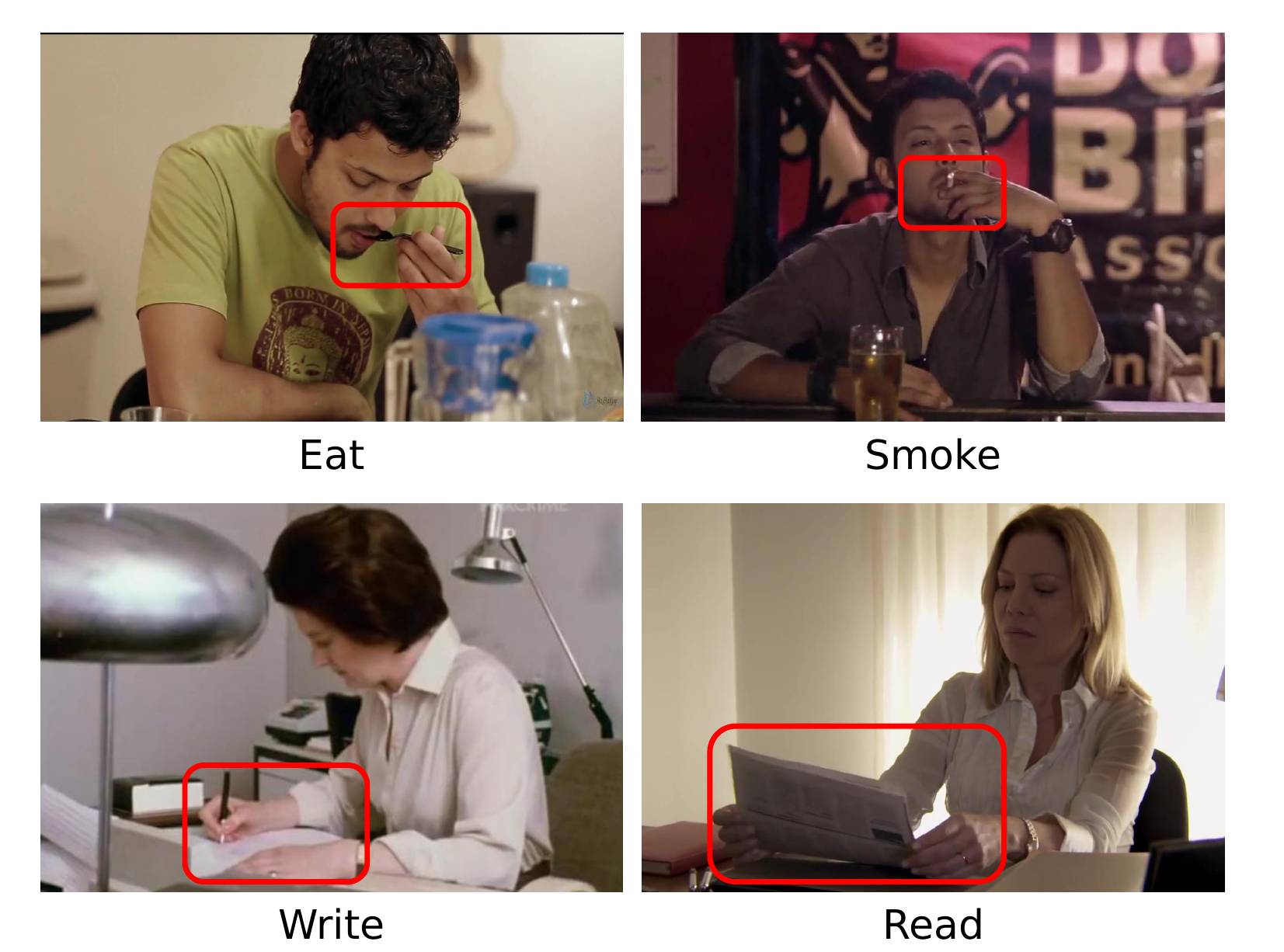}
	\end{center}
	\caption{
		Actor-centric action recognition heavily relies on local details (indicated by red rectangles) to distinguish between fine-grained actions.
		The top row shows the discriminative local region  between ``eat'' and ``smoke''  while the bottom row shows that between ``write'' and ``read''.}
	\label{fig:fig_introduction}
\end{figure}

This paper focuses on recognizing actor-centric actions for spatio-temporal action detection, including person pose actions, person-object interaction actions, and person-person interaction actions, which have wide applications across robotics, security, and health. The mainstream approaches \cite{cite_lfb, cite_slowfast, cite_action_transformer,
	cite_ava, cite_acrn} of action detection follow the Faster-RCNN~\cite{cite_faster_rcnn} style pipeline: classifying 3D convolutional features RoI-pooled over actor proposals into action categories.
Although they achieve significant progress, the action detection performance remains relatively low~\cite{cite_ava}.

RoI-Pooling based pipeline is the de facto standard for object recognition, which has been validated to be more efficient while without any accuracy degradation in Fast RCNN~\cite{cite_fastrcnn} and Faster RCNN~\cite{cite_faster_rcnn}.
Different from object recognition, actor-centric action recognition heavily depends on discriminative local regions to distinguish between fine-grained actions  as shown in Figure~\ref{fig:fig_introduction}. \eg, one differentiates ``eat'' from ``smoke'' via mouth regions of actors only. Therefore, actor-centric action recognition requires representations preserve more spatial details.
However, video backbone networks are memory-consuming and have to take low-resolution images as inputs.
This analysis prompts two questions:
\begin{itemize}
    \setlength{\itemsep}{0pt}
	\item Can RoI-Pooling based pipeline, which is widely used in the mainstream approaches of actor-centric action recognition, preserve the discriminative spatial details?
	\item If not, how to preserve adequate discriminative spatial details?
\end{itemize}

\begin{figure}
	\begin{center}
		\includegraphics[width=0.6\linewidth]{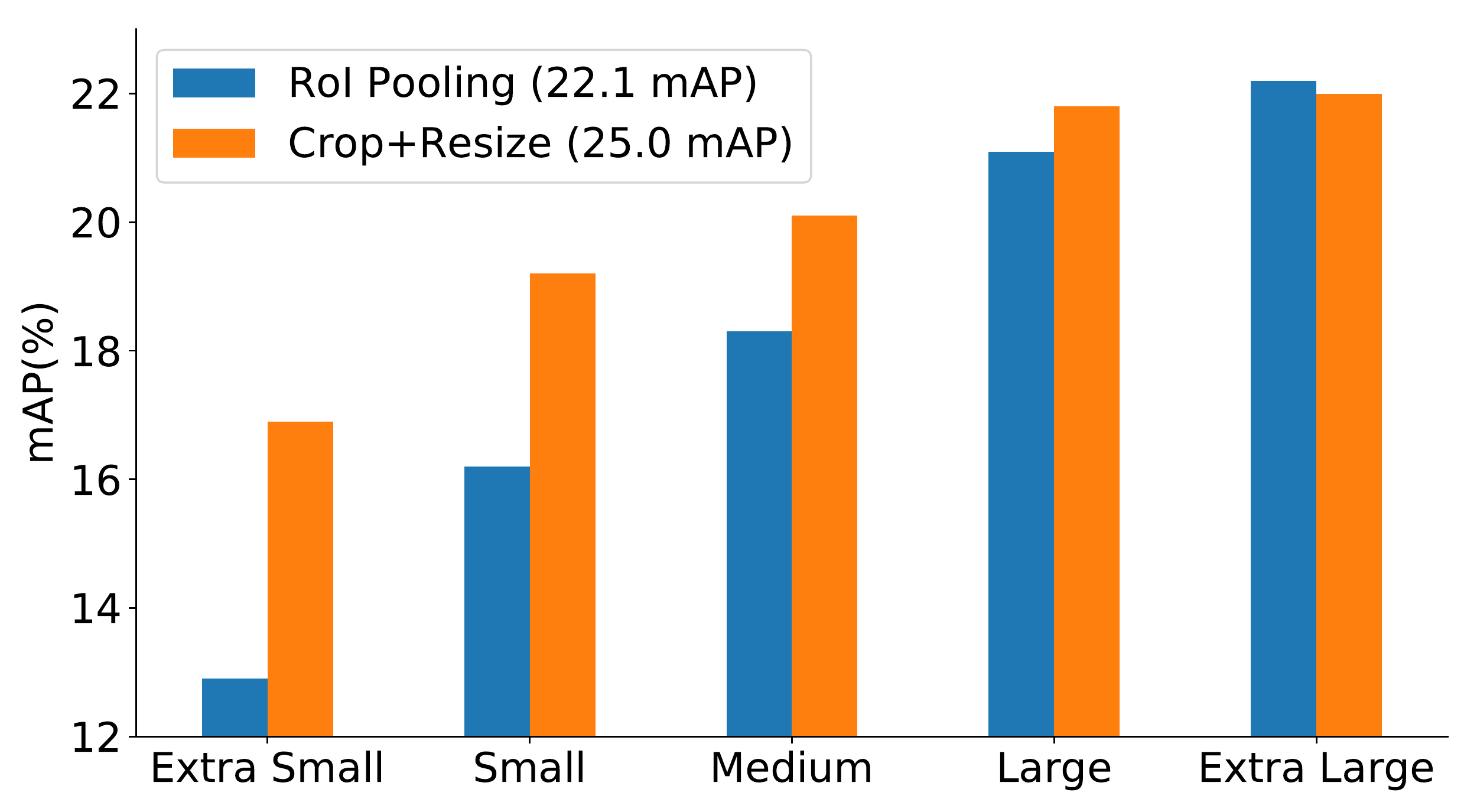}
	\end{center}
	
	\caption{
		We revisit RCNN-like method for action detection where actor boxes are cropped directly from original video and resized to a fixed resolution.
		Comparison of frame-level mAP performance at different bounding box sizes on AVA~\cite{cite_ava}}
	\label{fig:fig_introduction_comp}
\end{figure}

To answer the first question,  we conduct experiments with the representative RoI-Pooling based pipeline  on AVA dataset~\cite{cite_ava}, and evaluate action detection performance  with different actor box sizes. We empirically
found that RoI-Pooling based pipeline performs poorly when the bounding box size of actor is small, and action detection performance is highly correlated with actor size as shown in Figure~\ref{fig:fig_introduction_comp}.
We believe that the RoI-Pooling based pipeline losses many discriminative spatial details from bottom to top during network inference when actor sizes are small.

To answer the second question, we revisit RCNN-like method~\cite{cite_rcnn} in actor-centric action recognition.
Given actor bounding boxes predicted by an off-the-shelf person detector, we crop and then resize each actor to a fixed resolution, which is fed into an 3D CNN network to extract actor-centric features with full resolution for action classification.
In this way, our approach can enlarge the input resolution of small actors and preserve their discriminative spatial details, which suppresses the performance degradation due to small sizes of actors as shown in Figure~\ref{fig:fig_introduction_comp}.

Contextual information plays an important role in actor-centric action recognition, especially for actions related to person-person interactions or person-object interactions, such as ``talk to a person'' and ``ride a bicycle''.
To this end, we fuse the scene feature from the whole video clip and long-term feature from long-range temporal context, which further boosts the final performance.
This makes RCNN context-aware.

Although simple, proposed Context-Aware RCNN is remarkably effective. We extensively evaluate it on two popular action detection datasets, \ie, AVA~\cite{cite_ava} and  JHMDB~\cite{cite_jhmdb}.
It achieves new state-of-the-art results.
Specifically,  it with I3D ResNet-50 pushes the mAP of AVA to 28.0\%, improving that of the best ever reported method Long-Term Feature Bank (LFB)~\cite{cite_lfb} by 2.2\% and even outperforming LFB with a much bigger backbone I3D ResNet-101.

Our main \textbf{contributions} are summarized in two aspects:
\begin{itemize}
	\item We empirically investigate the drawback of current de facto standard pipeline of action detection and find that it losses many discriminative spatial details due to small resolutions of actors.
	\item We revisit RCNN-like method and propose a simple yet effective Context-Aware RCNN for action detection, which achieves a new state-of-the-art performance on two popular datasets. \ie, AVA~\cite{cite_ava} and JHMDB~\cite{cite_jhmdb}. Thanks to its simplicity, it is easy to implement and can serve as a strong baseline and start point for further research in the field of action detection.
\end{itemize}

\begin{figure*}[t]
	\begin{center}
		\includegraphics[width=1.0\linewidth]{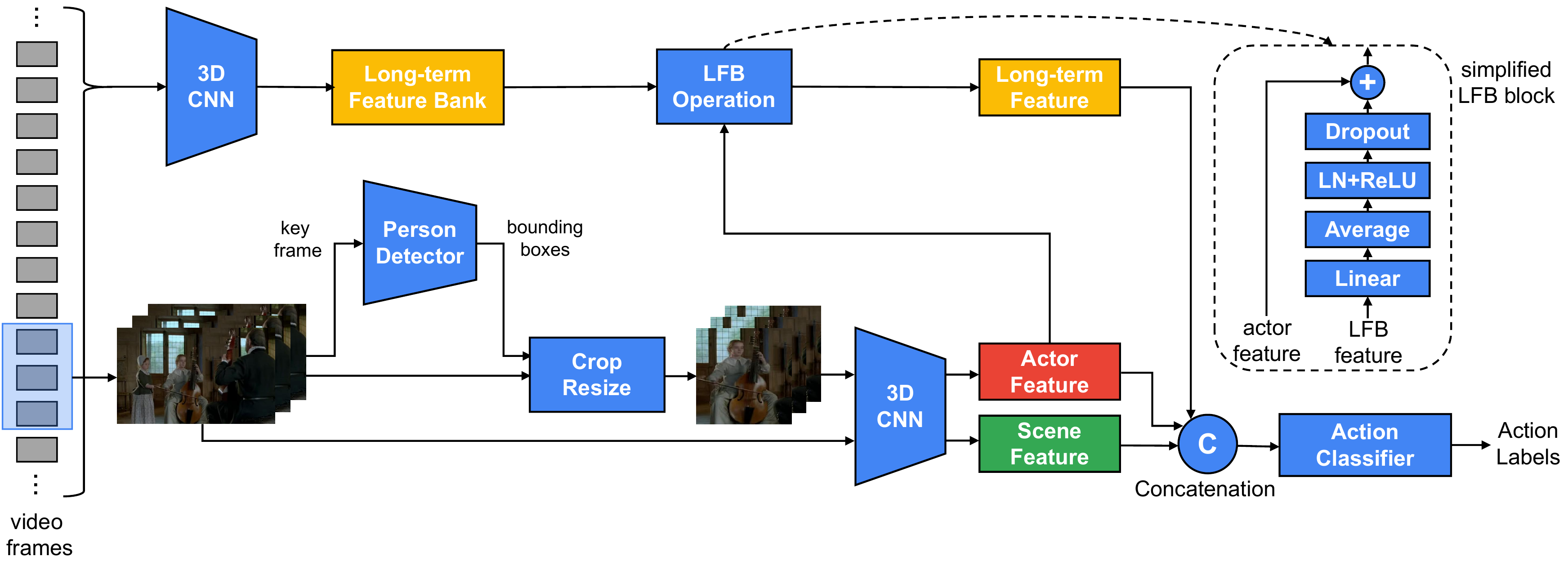}
	\end{center}
	\caption{
		An overview of proposed Context-Aware RCNN.
		We first localize actors at key frames using one person detector.
		Then, actors are cropped from the video clip, resized to one fixed resolution, and fed into one 3D CNN to extract actor features.
		Scene features and long-term features are also extracted as context information.
		Finally,  these three features are aggregated and fed into one action classifier to predict action labels
	}
	\label{fig:fig_framework}
\end{figure*}

\section{Related Work}
{\bf Action recognition.}
Thanks to the breakthrough of Convolutional Neural Networks (CNN)~\cite{cite_cnn, cite_alexnet, cite_vgg, cite_inception, cite_resnet}, video action recognition has evolved rapidly in recent years~\cite{cite_twostream,Wang0T15,GanWYYH15,cite_c3d,cite_nonlocal,cite_slowfast,cite_tsn}.
Many datasets~\cite{cite_hmdb, cite_ucf101, cite_youtube8m, cite_kinetics} have been proposed to foster research in video classification.
Simonyan~\etal~\cite{cite_twostream} designed a two-stream architecture to capture appearance and motion information with different network streams.
The two-stream CNNs achieved superior performance, but need to perform time-consuming optical flow calculation in advance.
3D CNNs~\cite{cite_c3d, cite_c3d2} extended 2D convolution to model static appearance and temporal motion directly from stacked RGB inputs.
Recent methods decomposed the 3D convolutions into separate 2D spatial and 1D temporal convolutions~\cite{cite_21d, cite_s3d, cite_p3d}.
There were also efforts that explore long-range temporal modeling in videos~\cite{cite_tsn, cite_longvideo,ZhangGHS020}.
Wang~\etal~\cite{cite_tsn} utilized a temporal segment network (TSN) to perform sparse sampling and temporal aggregation.
These works mainly focused on action recognition on well-trimmed videos, where the models only need to classify short video clips to action labels.
However, most videos are untrimmed and long in the practical applications.
Some recent works explored temporal action detection~\cite{cite_ssn, cite_bsn, cite_rethink_faster_rcnn, WangXLG17} and spatio-temporal action localization~\cite{cite_acrn, cite_lfb, cite_tcnn, cite_act, cite_ava, cite_action_transformer} on untrimmed videos.

{\bf Spatio-temporal action localization.}
Most action detection models~\cite{Wang0TG16, cite_acrn, cite_lfb, cite_tcnn, cite_act, cite_ava, cite_action_transformer,moc} extended object detection frameworks to handle videos.
Actor boxes were first computed by person detector and then classified to action labels.
Many of recent works focused on incorporating context information to improve recognizing human action~\cite{cite_acrn, cite_lfb, cite_action_transformer}.
ACRN~\cite{cite_acrn} computed pair-wise relation information from actor and global scene features, and generated relation features for action classification.
In ~\cite{cite_action_transformer}, a transformer architecture~\cite{cite_transformer} was used to aggregate features from the spatio-temporal context around the person.
Long-term feature bank~\cite{cite_lfb} was proposed to provide long-range information to video models.

{\bf Region-based convolutional neural networks.}
Our action detection framework is closely related to region-based object detectors~\cite{cite_rcnn, cite_fastrcnn, cite_faster_rcnn, cite_maskrcnn}.
RCNN~\cite{cite_rcnn} fed each Region of Interest (RoI) into convolutional networks independently to extract its corresponding feature vector.
SPPNet~\cite{cite_spp} and Fast RCNN~\cite{cite_fastrcnn} utilized RoI-Pooling to extract fixed length representation for each RoI from shared feature map.
Faster RCNN~\cite{cite_faster_rcnn} further improved efficiency of RCNN by using Region Proposal Network (RPN) to generate proposals.
Most recently, Cheng~\etal~\cite{cite_revisit_rcnn} found that redundant context information could lead to inferior classification capacity of Faster RCNN.
They used RCNN as a complement to improve the classification performance of Faster RCNN.
In this work, we adopt a RCNN-like network to extract actor features for actor-centric action classification.

\section{Approach}

For the task of spatio-temporal action localization, we need to localize every person at each key frame, and recognize their actions.
In this section, we will give a detailed description of our action detection approach.
First, we present an overview of proposed Context-Aware RCNN.
Then, we introduce how to extract discriminative features of each person using RCNN-like network~\cite{cite_rcnn}.
Finally, we take into account scene context and long-range context in our model.

\subsection{Method Overview}
The overall framework of our proposed network for  action detection is illustrated in Figure~\ref{fig:fig_framework}.
Our action detection framework follows the popular paradigm of frame-based action detection~\cite{cite_slowfast, cite_lfb, cite_acrn}, which contains two key stages: actor localization and action classification.
Given a key frame from the original video, a person detector is first used to localize actors at the key frame, obtaining a set of 2D bounding boxes.
Then for each actor bounding box, the local appearance features around actor are extracted for the classification of human activities.

Motivated by the success of 2D object detection algorithms, recent action detection methods~\cite{cite_slowfast, cite_lfb, cite_acrn} typically follow Faster-RCNN~\cite{cite_faster_rcnn} architecture to
use 3D RoI-Pooling to extract actor features from the clip feature map based on actor boxes.
In this work, we analyze that RoI-Pooling is sub-optimal to get discriminative features for action recognition because it losses many spatial details for small actors.
We use RCNN-like method to extract actor-centric features with full resolution.
Actor bounding boxes are used to crop actor tubes from the original video.
Then the cropped images are fed into non-local I3d network~\cite{cite_nonlocal} to extract deep representations of actors.

Moreover, we notice that context information is important to accurately recognize human activities.
Apart from using local actor features, we utilize a parameters shared network to extract global scene features of the whole video clip.
Besides that, a simplified non-local operation is used to capture long-range temporal context information from long-term feature bank~\cite{cite_lfb}.
Finally, actor features, global scene features and  long-term features are aggregated by concatenation for action classification.

\subsection{Extracting Actor Features}
%person detector
{\bf Person detector.}
For accurate actor localization, we follow previous works~\cite{cite_lfb, cite_slowfast} that use an off-the-shelf person detector to pre-compute person proposals.
The person detector takes the key frame as input, and outputs a set of act bounding boxes.
We use the proposals provided by~\cite{cite_lfb} on the AVA dataset to perform a fair comparison with state-of-the-art methods~\cite{cite_lfb, cite_slowfast}.
The proposals are detected by a Faster RCNN with a ResNeXt-101-FPN.
For JHMDB dataset, we train a Faster RCNN with a ResNet-50-FPN on its training set.

%backbone
{\bf Backbone.}
We use the I3D ResNet-50 network with non-local blocks as the backbone model for action classification.
The backbone model is pre-trained on ImageNet and 'inflated' into a 3D network using the I3D~\cite{cite_I3d} method.
Then the model is pre-trained for video classification on Kinetics-400~\cite{cite_I3d} equipped with non-local blocks~\cite{cite_nonlocal}.
Following the recommendation from~\cite{cite_lfb}, we set the stride of res$_5$ to 1 and use a dilation of 2 in res$_5$.
The resulting network downsamples the temporal dimension by a factor of $2$, and downsamples the spatial dimension by a factor of $16$.
Given an input of shape $T \times H \times W \times 3$, the backbone network outputs one feature map with a shape $T/2 \times H/16 \times W/16 \times 2048$.

{\bf Extracting actor features by RCNN.}
\label{section:actor_feature}
% RoI-pooling
% We first briefly review the RoI-Pooling operation in the Faster-RCNN detection framework.
RoI-Pooling~\cite{cite_fastrcnn} is introduced to extract Region of Interest (RoI) features from convolutional feature maps to a fixed size representation for the object classification.
RoIAlign~\cite{cite_maskrcnn} fixes the misalignments between the RoI and the extracted features by removing the harsh quantization of RoI-Pooling.
% drawbacks of RoI-pooling
However, the RoI-Pooling (or RoIAlign) may not be optimal to extract discriminative features for fine-grained action classification.
To correctly distinguish fine-grained action classes (\eg ``smoking'' and ``eating''), one has to focus on the local representative patterns (\eg ``cigarette'' and ``food'').
CNNs typically have a large downsample stride and receptive field at last layers of the network.
And state-of-the-art video models require dense sampling in time to achieve high recognition performance.
Therefore, to fit in GPU memory the video inputs to action detection framework must keep low-resolution (\eg  $224\times224$ in ~\cite{cite_slowfast, cite_lfb} and $400\times400$ in ~\cite{cite_action_transformer}).
For small actors, their sizes are very small at the last feature map of the backbone CNN network and local detail information is lost.
In such case, the current action detection architectures have poor sensitivity to finer details for fine-grained action recognition.

% advantages  of RCNN
Given the above analysis, we can conclude that using coarse representations from RoI-Pooling could not utilize the full potential of the classification power of deep video models.
In this work, we propose to use RCNN-like network in the action detection framework.
That is to say, actors are cropped directly from original video clip and resized to one fixed resolution.
In this way, the network can only see visual contents in their bounding boxes.
Moreover, the image inputs of small actors are enlarged to capture fine-grained details.

% box scale,  crop,  warp,  classification, base model
Our model takes the short video clip of $T$ frames as the network input, sparsely sampled from 64 neighboring frames of the key frame with a temporal stride $\tau$.
For an actor bounding box at key frame, we replicate the box along the temporal axis.
We conduct cropping according to the replicated box at each frame and resize the resulted image patches to a fixed resolution $H \times W$.
Then the actor clip with shape $T \times H \times W \times 3$ is fed to the backbone, followed by a global average pooling, resulting in a $2048$-dimensional feature vector as the actor feature.
For incorporating visual information around actor, we expand the actor bounding boxes slightly during training and inference.

\subsection{Context Modeling}
\label{section:context}
Context information plays an important role for understanding human activities,
Using only local actor features, the model could have poor performance due to the lack of context information.
We take into account two kinds of context information in our model to make RCNN context-aware: scene context in the short video clip and long-term context over the entire span of long video.

{\bf Scene context features.}
We feed the entire video clip to a 3D convolutional network, followed by global average pooling, to yield a 2048-dimensional scene feature vector.
A parameters shared backbone network is used to compute actor features and scene features.
We have tried to use independent backbone networks. However, it resulted in a poor performance.

{\bf Long-term context features.}
We adapt offline memory network architecture: long-term feature bank (LFB)~\cite{cite_lfb} to capture long-range temporal information.
We compute LFB as all actor features centered at the current clip within window size of $61$ seconds.
As illustrated in Figure~\ref{fig:fig_framework}, LFB operation, which consists of simplified LFB blocks, is used to extract long-term features  by taking LFB and actor features as inputs.
We pre-process these two kinds of input using dimension reduction and dropout following~\cite{cite_lfb}.
Different from~\cite{cite_lfb}, we replace the softmax attention weighted sum with average pooling and remove the last linear layer in the original LFB NL block,
Empirically, we found that this simplified version has a similar performance to original LFB NL block.
Three simplified LFB blocks are used as LFB operation in our experiments.
The output of LFB operation is a $512$-dimensional long-term feature vector.

% \subsection{Training Loss}
% % feature fusion + classification
Finally, actor features, scene features and long-term features are aggregated by concatenation and fed into the action classifier.
For AVA dataset, which is a multi-label classification task, we use the per-class sigmoid loss.
For JHMDB dataset, we use the softmax loss.

\section{Experiments}
In this section, we first introduce two widely-adopted datasets for spatio-temporal action localization and the implementation details of our approach.
Then, we perform a number of ablation studies to understand the effects of proposed components in our model.
We also compare the performance with the state-of-the-art methods to show the effectiveness and generality of our model.

\subsection{Datasets and Implementation Details}
{\bf Datasets.}
We conduct experiments on two publicly available spatio-temporal action localization datasets, \ie, the AVA~\cite{cite_ava} and the JHMDB~\cite{cite_jhmdb} dataset.

AVA~\cite{cite_ava} is a recently released large-scale action detection dataset.
We use the AVA version $2.1$ benchmark, which is composed of 211k training and 57k validation video segments.
Annotations are provided for key frames sparsely sampled at $1$ FPS.
Each person at key frame is labeled with one bounding box together with multi-label action labels from 80 atomic action classes.
Following the standard protocol~\cite{cite_ava}, we evaluate over 60 classes.
We report mean Average Precision (mAP) performance on frame level using an IoU threshold of 0.5.

JHMDB~\cite{cite_jhmdb} dataset contains 928 temporally trimmed short video clips with 21 action classes.
Every frame in JHMDB is annotated with one actor bounding box and a single action label.
As is standard practice, we conduct experiments on three training/validation splits, and report the average frame-level mAP with IoU threshold of 0.5 over three splits.

\begin{table}[!p]
\begin{center}
    \caption{AVA validation results using only actor features. We show the mAP performance on frame level using IoU threshold of 0.5}
	
	\setlength{\tabcolsep}{6mm}{
		\begin{subtable}{1.0\textwidth}
			\begin{center}
			    \caption{ {\bf RoI Pooling vs. RCNN-like method.}
				RCNN-like method is consistently better than the RoI Pooling with different input sampling $T \times \tau$.
				$T$ denotes the number of input frames, and $\tau$ denotes temporal sample stride}
				\begin{tabular}{c|c|c}
					\hline
					Method                                         & $T \times \tau$    & mAP  \\
					\hline
					\multirow{3}{*}{RoI Pooling}                       & $8 \times 8$  & 20.1 \\% \cline{2-3}
					& $16 \times 4$ & 21.9 \\ %\cline{2-3}
					& $32 \times 2$ & 22.1 \\
					\hline
					\multirow{3}{*}{Crop+Resize}                          & $8 \times 8$  & 23.1 \\
					& $16 \times 4$ & 24.7 \\
					& $32 \times 2$ & 25.0 \\
					\hline
				\end{tabular}
				\label{table:table_1a}
			\end{center}
			\label{}
		\end{subtable}
	}
	
	\setlength{\tabcolsep}{0.5mm}{
		\begin{subtable}{1.0\textwidth}
			\begin{center}
			    \caption{ {\bf Performance comparison with different actor box sizes.}
				Extracting actor features via cropping and resizing is helpful to improve the  performance on small actor boxes
				}
				\begin{tabular}{ c|c|c|c|c|c}
					
					\hline
					\diagbox[width=5em,trim=l]{Method}{Size}  & Extra small & Small & Medium & Large &  Extra large  \\
					\hline
					RoI Pooling         &  $12.9$  & $16.2$ & $18.3$ & $21.1$ & $22.2$          \\
					Crop+Resize            &  $16.9$  & $19.2$ & $20.1$ & $21.8$ & $22.0$          \\
					\hline
					Improvement            &  $+4.0$  & $+3.0$ & $+1.8$ & $+0.7$ & $-0.2$          \\
					\hline
					
				\end{tabular}
				\label{table:table_1b}
			\end{center}
			\label{}
		\end{subtable}
	}
	
	\setlength{\tabcolsep}{0.5mm}{
		\begin{subtable}{1.0\textwidth}
			\begin{center}
			    \caption{ {\bf Performance of different input resolutions using the cropping with resizing method.}
				Input resolution is crucial for actor-centric action recognition
				}
				\begin{tabular}{ c|c|c|c|c }
					
					\hline
					Resolution  &  $224 \times 224$  &  $192 \times 192$  &  $160 \times 160$  &  $112 \times 112$   \\
					\hline
					mAP         &  $25.0$  &  $24.3$  &  $23.7$  &  $21.4$          \\
					\hline
					
				\end{tabular}
				\label{table:table_1c}
			\end{center}
			\label{}
		\end{subtable}
	}
	
	\setlength{\tabcolsep}{1mm}{
		\begin{subtable}{1.0\textwidth}
			\begin{center}
			    \caption{ {\bf Performance comparison with different actor numbers in the frame.}
				The performance decreases as more people appear in the key frame
				}
				\begin{tabular}{ c|c|c|c|c|c|c }
					
					\hline
					\diagbox[width=5em,trim=l]{Method}{Count}  &  $[1,1]$  &  $[2,3]$  &  $[4,5]$  &  $[6,7]$  &  $[8,9]$  &  $[9,37]$     \\
					\hline
					RoI Pooling         &  $29.2$  &  $23.7$  &  $25.3$  &  $25.1$  &  $22.4$  &  $17.2$                 \\
					Crop+Resize            &  $35.2$  &  $25.9$  &  $27.1$  &  $27.6$  &  $26.3$  &  $22.9$                 \\
					\hline
					Improvement            &  $+6.0$  & $+2.2$ & $+1.8$ & $+2.5$ & $+3.9$ & $+5.7$          \\
					\hline
					
				\end{tabular}
				\label{table:table_1d}
			\end{center}
			\label{}
		\end{subtable}
	}
	
	\setlength{\tabcolsep}{2mm}{
		\begin{subtable}{1.0\textwidth}
			\begin{center}
			    \caption{ {\bf Performance comparison when expanding actor bounding boxes by different scales.}
				Here ``$^*$'' indicates using RoI Pooling. The others use cropping with resizing
				}
				\begin{tabular}{ c|c|c|c|c|c|c }
					
					\hline
					Max scale  &  $1.2$  &  $1.5$  &  $1.8$  &  $2.0$  &  $2.5$ & $1.5^*$  \\
					\hline
					mAP        &  $24.1$  &  $25.0$  &  $24.8$  &  $25.0$  &  $24.3$  & $21.6$        \\
					\hline
					
				\end{tabular}
				\label{table:table_1e}
			\end{center}
			\label{}
		\end{subtable}
	}
	
	\label{table:table_1}
\end{center}
\end{table}

{\bf Implementation details.}
We adopt the synchronous SGD with a mini-batch size of 16 on 8 GPUs.
We keep batch normalization layers frozen during training.
For the AVA dataset,
we train the network for 140k iterations with a learning rate of 0.04 which is decreased by a factor of 10 at iteration 100k and 120k.
We use linear warmup for the first 1000 iteration.
Dropout of 0.3 is used before the final classifier layer and weight decay is set as $10^{-6}$.
For the JHMDB dataset,
we adopt the same learning rate schedule except for using an initial learning rate of $0.001$.
Dropout of 0.5 and weight decay of $10^{-7}$ are used.

Data augmentation is used to improve generalization.
We randomly extend boxes in height and width with scale  $\in[1, s]$ during training and use a fixed scale  $s$ during inference.
The default $s$ is $1.5$.
It is then cropped from the image and resized to $224 \times 224$ for training and $256 \times 256$ for testing.
If a box crosses the image boundary, we crop the region within the image.
For the entire video frame inputs, we perform random scaling such that the short side $\in [256, 320]$, and random cropping of $224 \times 224$ for training.
At test time, we rescale the short side to $256$ pixels and use a single center crop of $256 \times 256$.
Moreover, random flipping is used during training.

We initialize the 3D ResNet-50 backbone network with non-local blocks by pre-training on action classification dataset Kinetics-400~\cite{cite_kinetics}.
The model using only actor feature and scene feature can be trained end-to-end.
We follow~\cite{cite_lfb} to train the model with long-term feature bank in two stages:
we first fine-tune the 3D CNN to compute features for long-term feature bank.
Then we fine-tune the whole model which aggregates long-term features and short-term features.

\subsection{Ablation Study}
In this subsection, we perform detailed ablation studies on the AVA dataset
to investigate effects of different components in our model.
This dataset provides
a challenging benchmark for our analysis as it consists of a
large set of examples.

{\bf RoI-Pooling v.s Crop+Resize.}
We begin our experiments by comparing RCNN-like method for extracting actor features with RoI-Pooling based method.
To this end, we train the model only using actor features for action classification without using scene feature or long term feature.
For RoI-Pooling baseline, following state-of-the-art methods~\cite{cite_slowfast, cite_lfb}, we first average pool the last feature map of res$_5$ over the time axis. Then RoIAlign~\cite{cite_maskrcnn} is used, followed by spatial max pooling, to yield a $2048$-dimensional vector as actor features.
The results are listed in Table~\ref{table:table_1a} at different input frames.
We first observe that taking more frames as inputs gives steady performance improvement, due to incorporating more temporal information.
Moreover, using cropped and resized images as inputs brings a significant performance boost over the RoI-Pooling method across all temporal lengths.
In particular, it achieves mAP of $25.0\%$ only using non-local 3D ResNet-50 to extract actor features for action classification without context modeling.
It outperforms RoI-Pooling method by $2.9\%$.
We further perform a number of ablation experiments to better understand where does the gain come from.
By default, $T \times \tau = 32 \times 2$ (sample 32-frame clip with temporal stride 2 as inputs) and cropping with resizing are used .

{\bf Impact of box size.}
Table~\ref{table:table_1b} compares the performance with respect to the size of the actor bounding box.
Following~\cite{cite_action_transformer}, we break down the performance of model into $5$ bins according to the size of actor bounding box: ``extra small'', ``small'', ``medium'', ``large'' and ``extra large''.
Each bin is defined by the thresholds of percentage image area covered by the box:
 $(0, 8.11\%]$ for ``extra small'',
 $(8.11\%, 17.11\%]$ for ``small'',
 $(17.11\%, 29.24\%]$ for ``medium'',
 $(29.24\%, 47.2\%]$ for  ``large'',
 and $(47.2\%, 100.0\%]$ for  ``extra large''.
And every bin has similar ground truth box count on validation set.
We split the prediction and ground truth boxes on validation set into bins, and evaluate mAP at each bin.
We can see that bigger actor bounding boxes obtain better performance.
The models perform poorly when actor boxes are small due to the lost of local detail information.
Moreover, we find extracting actor features via cropping and resizing is helpful to improve the performance on small actor boxes, and offers similar performance on big actor boxes.
This demonstrates that RoI-Pooling method losses discriminative spatial details of actors due to small resolution of actors in the RoI-Pooling layer.
And RCNN-like method enlarges the actor input size which can help better preserve finer detail information.
We believe this suggests a promising future research
direction that extracts high-resolution representations for action recognition in action detection systems.

{\bf Impact of input resolution.}
We further investigate the effect of the actor input resolution.
The actor boxes are cropped from the images and resized to different resolutions as network inputs for actor feature extraction.
The results are reported in Table~\ref{table:table_1c}.
As expected, it shows that using low resolution images as inputs results in the performance drops dramatically.
This confirms that input resolution is crucial for actor-centric action recognition.

\begin{figure*}[!t]
	\begin{center}
		\includegraphics[width=1.0\linewidth]{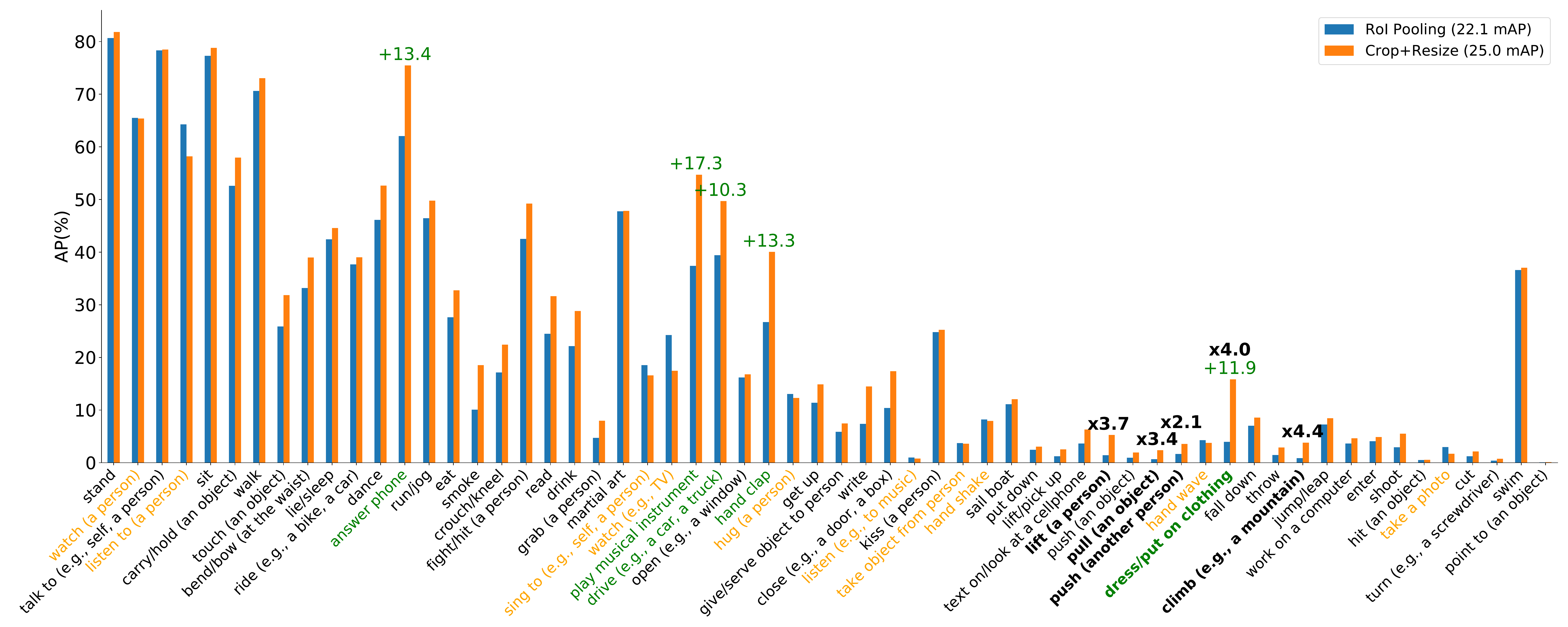}
	\end{center}
	\caption{
		Per-category AP on the AVA dataset.
		RoI-Pooling based method v.s RCNN-like method using cropping with resizing.
		Categories are sorted by the the number of examples.
		(\textcolor[rgb]{0,0.394,0}{Green}: 5 categories with largest absolute performance increase.
		 \textbf{Bold}: 5 categories with largest relative performance increase.
		 \textcolor[rgb]{0.785,0.519,0.05}{Orange}: categories with decreased performance)
	}
	\label{fig:fig_ap}
\end{figure*}

{\bf Impact of actor count in the scene.}
In Table~\ref{table:table_1d}, we compare the performance with respect to the number of GT boxes in the key frame.
The performance generally decreases as more people appear in the key frame.
The RCNN-like method achieves significant improvements when there is only one person in the frame.
We conjecture that single person scene relies more on local details for activity recognition.
Moreover the performance improvement is also large when the scene has very many people where the actor boxes are often small.

{\bf How does performance change across actions.}
Next, we compare the AP performance of RoI-Pooling baseline and RCNN-like method for each action class in Figure~\ref{fig:fig_ap}.
The action classes are sorted by the number of the examples.
The RCNN-like method improves performance in 50 out of 60 classes.
We highlight the 5 highest absolute performance increase categories and 5 largest relative increase categories.
Recognizing these action labels correctly needs to focus on the local representative patterns around the actor (\eg ``musical instrument'' for ``play musical instrument'').
RCNN-like method enlarges the resolution of small box and is able to focus on the local discriminative regions of the videos.
The RCNN-like method yields slightly worse performance on 10 classes:
``watch (\eg TV)'' ($-6.8$ AP), ``listen to'' ($-6.1$ AP), ``sing to'' ($-1.9$ AP) and so on.
These are categories where modeling context information is critical.
We conjecture that RoI-Pooling can incorporate scene information due to bigger receptive field.
But RCNN-like method can only extract actor feature within the box.
Such issues motivate us to  explicitly compute context features in our framework.

{\bf Impact of the scale to expand bounding box.}
%For incorporating visual information around actor and data augmentation, we random extend the box in height and width with scale  $\in[1, s]$ during training and use a fixed scale  $s$ during inference.
We experiment with different scales to expand the actor box.
The results are shown in Table~\ref{table:table_1e}.
Expanding the bounding box slightly is helpful to improve performance of our model.
However, too large scale hurts the performance due to incorporate redundant context information.
We also conduct experiment based on RoI-Pooling, which shows it doesn't help to improve RoI-Pooling method ($22.1\% \rightarrow 21.6\%$).
By default, the scale $1.5$ is used in the experiments.

\setlength{\tabcolsep}{4pt}
\begin{table}[!t]
\begin{center}
\caption{AVA validation results using scene features and long-term features.
We compare the performance of different methods to fuse context features.
NL denotes Non-Local.
Simple fusion methods (Concat, NL average) offer similar performance to complicated fusion methods
}

    \begin{tabular}{ c|c|c }

    \hline
      Scene feature    &  Long-term feature       &  mAP  \\
    \hline
      -   & -     &   $24.7$    \\
       \hline
      Concat  & -     &  $25.7$     \\
     Transformer+concat  & -     &  $25.8$     \\
       \hline
      Concat  & NL attention   &  $27.6$  \\
      Concat  & NL average    &  $27.8$  \\
      Concat  & NL average w/o last linear    &  $28.0$  \\
    \hline
	
    \end{tabular}

\label{table:table_2}
% \vspace{-10mm}

\end{center}
\end{table}
\setlength{\tabcolsep}{1.4pt}

\begin{table}[!t]
\begin{center}
\caption{Comparison with the state-of-the-art methods on the AVA and JHMDB datasets}

\setlength{\tabcolsep}{0.75mm}{
\begin{subtable}{1.0\textwidth}
    \begin{center}
    \caption{ Comparison with  state-of-the-art on the AVA v2.1}
    \begin{tabular}{ c|c|c|c|c }

    \hline
      Method & Flow & Video Pretrain & Backbone &  mAP  \\
    \hline
       AVA baseline~\cite{cite_ava} &\checkmark& Kinetics-400 & I3D &  $ 15.6$    \\
       ACRN~\cite{cite_acrn} &\checkmark& Kinetics-400 & S3D &  $17.4$    \\
       Relation Graph~\cite{cite_relation_graph} & &Kinetics-400&R50-NL&  $22.2$    \\
       VAT~\cite{cite_action_transformer}  & & Kinetics-400 & I3D &  $ 25.0 $    \\
       SlowFast~\cite{cite_slowfast}  & & Kinetics-400  & R50 &  $24.2$    \\
       SlowFast~\cite{cite_slowfast}  & & Kinetics-400 & R101 &  $26.3$    \\
       SlowFast~\cite{cite_slowfast}  & & Kinetics-600 & R101-NL &  $ 28.2 $    \\
       LFB~\cite{cite_lfb} & & Kinetics-400 & R50-NL &  $25.8$    \\
       LFB~\cite{cite_lfb} & & Kinetics-400 & R101-NL &  $27.1$    \\
    \hline
      Context-Aware RCNN & &Kinetics-400& R50-NL &  $28.0$  \\
    \hline

    \end{tabular}
    \label{table:table_3a}
    \end{center}
    \label{}
\end{subtable}
}

\setlength{\tabcolsep}{0.1mm}{
\begin{subtable}{1.0\textwidth}
    \begin{center}
    \caption{ Comparison with  state-of-the-art on the JHMDB dataset}
    \begin{tabular}{ c|c|c|c|c }
    	
    	\hline
    	Method & Flow & Video Pretrain & Backbone &  mAP  \\
    	\hline
    	Two-stream RCNN~\cite{cite_two_stream_rcnn} &\checkmark&   & VGG &  $58.5$    \\
    	T-CNN~\cite{cite_tcnn} & &   & C3D &  $61.3$    \\
    	ACT~\cite{cite_act} &\checkmark&  & VGG &  $65.7$    \\
    	AVA baseline~\cite{cite_ava} &\checkmark& Kinetics-400 & I3D &  $73.3$    \\
    	ACRN~\cite{cite_acrn} &\checkmark& Kinetics-400 & S3D &  $77.9$    \\
    	\hline
    	Context-Aware RCNN & &Kinetics-400& R50-NL &  $79.2$  \\
    	\hline

    \end{tabular}
    \label{table:table_3b}
    \end{center}
    \label{}
\end{subtable}
}
\label{table:table_3}
% \vspace{-10mm}
\end{center}
\end{table}

{\bf Trade-off between performance and time efficiency.}
We measured the inference wall-clock time of RoI-Pooling based method and RCNN-like method on the validation set of the AVA dataset.
The result is as follows: RoI-Pooling based method achieves mAP $22.1\%$ using $4548$ seconds.
And RCNN-like method achieves mAP $25.0\%$ using $7500$ seconds.
RCNN-like pipeline uses around $1.65 \times$ more inference-time than RoI-Pooling based pipeline.
RCNN-like pipeline achieves better recognition performance at the cost of more inference-time.

{\bf Impact of context modeling.}
We now extend our model to use scene features and long-term features as context information.
The results are shown in Table~\ref{table:table_2}.
We first study the effect of scene features which are extracted from the entire video clip.
To reduce computational cost, we set $T \times \tau = 16 \times 4$ when scene features are used.
We concatenate the scene features and actor features for action classification.
We observe that the usage of scene features  bring significant performance improvement ($24.7\% \rightarrow 25.7\%$).
We also try to use actor features to attend to scene features in global feature map by  Transformer-styled blocks~\cite{cite_action_transformer}, but fail to find they offer obvious performance improvement.
We further add long-term features to our model, which are extracted by non-local function from long-term feature bank.
Different from~\cite{cite_lfb}, we use a simplified non-local version as described in Section~\ref{section:context}.
We can see that incorporating long-term features leads
 to large performance gain ($25.7\% \rightarrow 28.0\%$).
 Moreover, our simplified non-local version can achieve slightly better results.
 Interestingly, using simple average pooling offers similar performance to using attention weighted sum.

\begin{figure*}[t]
	\begin{center}
		\includegraphics[width=1.02\linewidth]{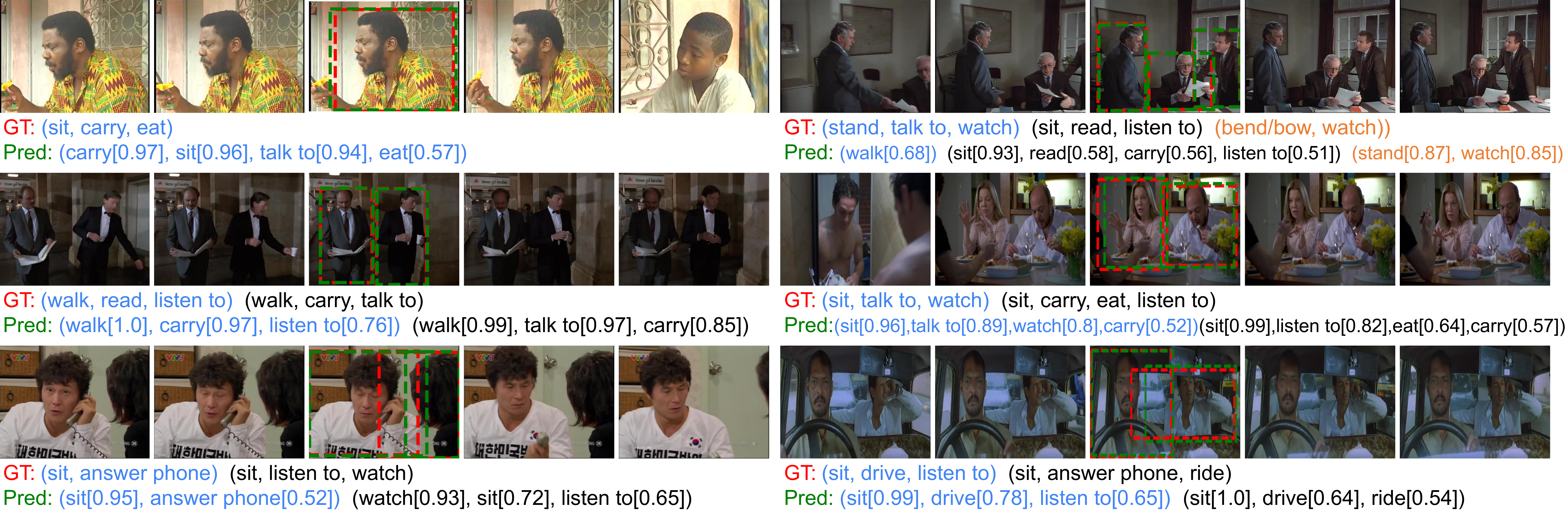}
	\end{center}
	\caption{
		Visualization on the AVA dataset.
		For each example, we show the key frame with ground-truth boxes (\textcolor[rgb]{1.0,0,0}{Red}) and predicted boxes (\textcolor[rgb]{0,0.394,0}{Green}),
		together with surrounding 4 frames.
		The ground-truth and predicted labels are presented under the pictures.
		Multiple labels of one actor are marked with same text color, and actors are labeled from left to right.
		Our best model with mAP $28.0$ is used
	}
	\label{fig:fig_visual}
% 	\vspace{-5mm}
\end{figure*}

\subsection{Comparison with the state of the art}
We compare our model with the state-of-the-art methods in Table~\ref{table:table_3}.
For fair comparison, we consider methods that only use single model and single crop for testing.
We also list other important factors such as backbone architectures.

Table~\ref{table:table_3a} shows the comparison with previous results on the AVA dataset.
Under the similar settings, which use Kinetics-400 to pre-train and ResNet-50 as backbone,  our method outperforms LFB~\cite{cite_lfb} by $2.2\%$ (28.0\% versus 25.8\%).
Our ResNet-50 model even yields better performance than LFB using deeper ResNet-101 (28.0\% versus 27.1\%).
Meanwhile, our model achieves comparable performance (28.0\% versus 28.2\%) to SlowFastNet~\cite{cite_slowfast} which use stronger backbone (SlowFast-ResNet-101) and more pre-training data (Kinetics-600).

We further evaluate our model on the JHMDB dataset.
The results and comparison with previous methods are listed in Table~\ref{table:table_3b}.
We only use actor features and scene features for action classification, because the videos of JHMDB dataset are short.
And $T \times \tau=16 \times 4$ is used.
Our method achieves the state-of-the-art performance of $79.2 \%$ by using only RGB frames.
This outstanding performance shows the effectiveness and generality of our model.

\subsection{Qualitative Results}

Finally, we qualitatively visualize several example predictions on the AVA dataset in Figure~\ref{fig:fig_visual}.
As we can see, the difficulty in the AVA lies in actor-centric action recognition, and the actor localization is less challenging~\cite{cite_ava}.
Overall, our model offers good action detection results, but still has difficulty to recognize action requiring fine grained discrimination.
For the example in the bottom right corner, the model fails to predict the right person's answering phone action.

\section{Conclusion}
In this paper, we have revisited RCNN-like method to extract actor features via cropping and resizing in action detection framework.
This approach provides a simple way to enlarge the input resolution of small actor boxes and preserve their discriminative spatial details.
% Experiments show that it offers significant performance improvement compared with RoI-Pooling method, especially on small actor bounding boxes.
% This demonstrates high-resolution information is critical for actor-centric action recognition.
Moreover, we present a simple and effective Context-Aware RCNN for action detection which aggregates actor features, scene features and long-term features.
It achieves state-of-the-art performance on challenging benchmarks.
We conduct comprehensive ablation studies to validate the effectiveness of different components in our model.
Our approach can serve as a simple and strong baseline for action detection.
In the future, we plan to further study how to improve the sensitivity of action detection model to fine-grained details of human activities.
Moreover, effective context reasoning method is also a promising research direction. \\

\noindent {\bf Acknowledgements.} This work is supported by SenseTime Research Fund for Young Scholars, the National Science Foundation of China (No. 61921006), Program for Innovative Talents and Entrepreneur in Jiangsu Province, and Collaborative Innovation Center of Novel Software Technology and Industrialization.

\clearpage

% ---- Bibliography ----
%
% BibTeX users should specify bibliography style 'splncs04'.
% References will then be sorted and formatted in the correct style.
%
\bibliographystyle{splncs04}
\bibliography{egbib}
\end{document}